\documentclass{article}

\PassOptionsToPackage{numbers, compress}{natbib}

\usepackage[final]{neurips_2023}

\usepackage[utf8]{inputenc} 
\usepackage[T1]{fontenc}    
\usepackage{hyperref}       
\usepackage{url}            
\usepackage{booktabs}       
\usepackage{amsfonts}       
\usepackage{nicefrac}       
\usepackage{microtype}      
\usepackage{xcolor}         
\usepackage{amsmath}
\usepackage{amssymb}
\usepackage{multirow}
\usepackage{graphicx}
\usepackage{natbib}
\usepackage{comment}

\title{Multi-Subject Personalization} 

\vspace{-2mm}
\author{
  Arushi Jain, Shubham Paliwal, Monika Sharma, Vikram Jamwal, Lovekesh Vig \\
  TCS Research, New Delhi, India\\
  \texttt{\{j.arushi, shubham.p3, monika.sharma1, vikram.jamwal,  lovekesh.vig\}@tcs.com} \\
}

\begin{document}

\maketitle

\begin{abstract}
Creative story illustration requires a consistent interplay of multiple characters or objects. However, conventional text-to-image models face significant challenges while producing images featuring multiple personalized subjects. For example, they distort the subject rendering, or the text descriptions fail to render coherent subject interactions. We present Multi-Subject Personalization (MSP) to alleviate some of these challenges. We implement MSP using Stable Diffusion and assess our approach against other text-to-image models, showcasing its consistent generation of good-quality images representing intended subjects and interactions.


\end{abstract}


\noindent \textbf{1. Introduction} 
Text-to-image generation deep models \cite{stable-diffusion, dalle, imagen} have found applications across various industries, delivering impressive visual assets for purposes such as advertising, marketing, entertainment, and creative content creation like story visualization. Fine-tuning through models such as LoRA \cite{lora}, Textual Inversion~\cite{textual-inversion}, and Dreambooth~\cite{dreambooth} further allows personalization of style and objects. Nevertheless, when generating more than one personalized subject, they frequently produce images with unclear or hybrid characteristics of multiple subjects (see Appx \ref{ap-sec:problem} Fig. \ref{fig:comp-results}). Recently, efforts have been made to address this limitation, with models like Custom-Diffusion~\cite{custom-diffusion} and Subject-Diffusion~\cite{subject-diffusion}, allowing for the generation of images with multiple personalized subjects. Nonetheless, these models still struggle to generate complex compositions with more than two customized subjects. To address these constraints, we present a novel approach for generating images featuring multiple personalized subjects based on Stable Diffusion~\cite{stable-diffusion}, including humans, animals, and objects. Our method enables more precise control over the subject's appearance and positioning within the generated image, thus offering artists the enhanced ability to craft complex scene imagery tailored to specific requirements.

\begin{figure*}[]
  \centering
  \includegraphics[width=\textwidth]{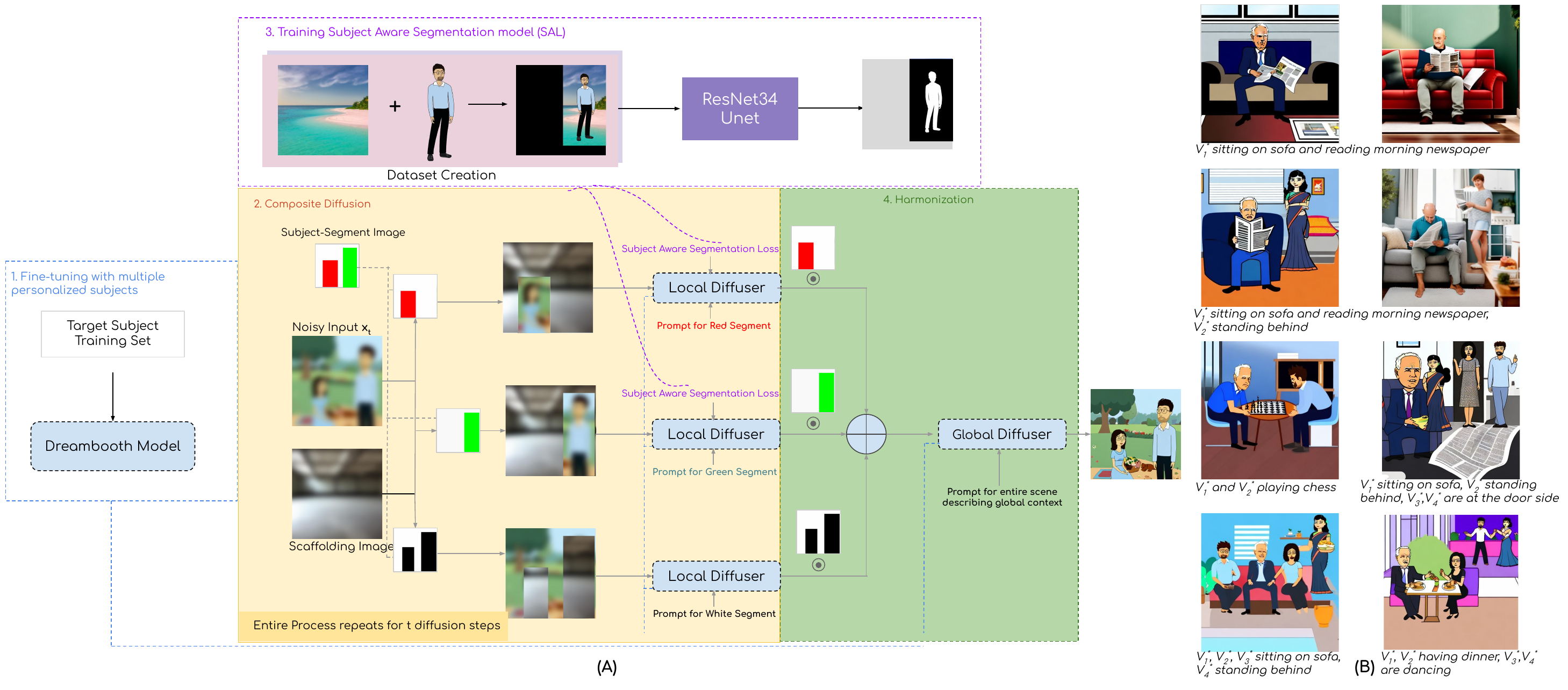}
  \vspace{-2mm}
  \caption{\small{(A) Overview of the MSP-Diffusion architecture, (B) Sample generations from the method}}
  \vspace{-2mm}
  \label{fig:msp-diffusion-architecture}
\end{figure*}

\noindent \textbf{2. Proposed Method: MSP-Diffusion} 
Our approach for Multi-subject Personalization, using diffusion-based image generation models, as shown in Fig.~\ref{fig:msp-diffusion-architecture}, has the following four salient features:

\noindent \textit{(i) Fine-tuning the model for multiple personalized subjects}: Our approach centers on embedding a specific subject instance within the output domain of a text-to-image diffusion model by associating each subject with a unique identifier. We achieve this by jointly fine-tuning the Dreambooth model using a limited set of sample images for each subject, represented by distinct modifier tokens, $V^{\ast}$, initialized with different rarely-occurring tokens. 
We incorporate training images featuring subjects in diverse angles and poses to enhance subject comprehension and, consequently, improve reverse diffusion-based generation (see Appendix \ref{ap-sec:dataset}). 

\noindent \textit{(ii) Controlling the positioning and orientation of multiple subjects}: We employ the Composite Diffusion technique~\cite{composite-diffusion}, as shown in Figure~\ref{fig:msp-diffusion-architecture}, to precisely control the spatial placement of personalized subjects using subject-segment layout image in the diffusion process (see Appendix Figure~\ref{fig:input-sample}). 
The approach compels the diffusion process to create an appropriate subject in the corresponding segment region, effectively resolving issues related to missing and hybrid subjects while ensuring the correct number of subjects are generated. We further provide an implementation involving ControlNet~\cite{control-net}, where we introduce {control conditioning inputs in the form of Openpose\cite{openpose} images}, 
in addition to text-based conditioning. The subject-segment image defines individual subjects, while the Openpose information controls the subject's posture and scale.

\noindent \textit{(iii) Ensuring precise subject-appearance}: In order to gain control over the desired subject's appearance in the generated image, we introduce a {subject-aware segmentation loss} (SAL) derived from a ResNet-34 based U-Net~\cite{unet} segmentation model trained on the same dataset of target subjects as used to fine-tune the Dreambooth model (see Appendix \ref{ap-sec:sal}). 

\noindent \textit{(iv) Harmonization of multiple subjects}: Similar to Composite Diffusion \cite{composite-diffusion}, we use a two step generation process. During the initial $k$ steps, when subjects in individual segments are constructed independently, the resulting composite often lacks harmony and smooth blending at the segment edges. To address this, harmonization is performed that allows each segment to develop in the context of the others. Additionally, we suggest the use of a global diffuser to capture the image's overall context, ensuring harmonization among the various segments containing personalized subjects.  


\noindent \textbf{3. Evaluation} 
We evaluate MSP-Diffusion on our self-compiled Subject-Dataset11 featuring $11$ diverse target subjects, including comic characters (male, female), human characters (male, female), pets, and still objects. We used three evaluation metrics for performance comparison: (CLIP-I)~\cite{textual-inversion}, (CLIP-T)~\cite{text-alignment}, and image alignment using our subject-aware similarity (SAS). 
Table~\ref{tab:comp-tab} shows that as the number of personalized subjects increases, Dreambooth, Textual Inversion and Custom-Diffusion exhibit reduced image alignment, whereas MSP-Diffusion without scaffolding image, achieves a CLIP-I score of 0.93.
Notably, our method consistently outperforms previous models in terms of CLIP-I, CLIP-T, and SAS scores when dealing with scenarios involving more than one personalized subject. Next, in the case of text alignment, the performance of our method with scaffolding image deteriorates as the number of subjects increases due to the generation of hybrid or missing subjects. Therefore, ControlNet~\cite{control-net} based Openpose is incorporated in the MSP-Diffusion which successfully mitigates the issue, albeit introducing a dependency on Openpose. Please refer to the Appendix~\ref{ap-sec:eval-metric} and  \ref{ap-sec:results} for more details and qualitative results.


\vspace{-2mm}
\begin{table}[!hbtp]
\centering
\caption{\small{Comparison of proposed MSP-Diffusion on Subject-Dataset11. Here, n, GD, SAL, and SAS refer to the number of personalized subjects, Global Diffuser, Subject-aware loss, and Subject-aware Similarity, respectively.}}
\vspace{-2mm}
\resizebox{\textwidth}{!}{%
\begin{tabular}{|c|c|ccc|ccc|ccc|ccc|}
\hline
\multirow{2}{*}{\textbf{Methods}} &
  \textbf{Configuration} &
  \multicolumn{3}{c|}{\textbf{ n = 1}} &
  \multicolumn{3}{c|}{\textbf{ n = 2}} &
  \multicolumn{3}{c|}{\textbf{ n = 3}} &
  \multicolumn{3}{c|}{ \textbf{ n \textgreater 3}} \\ \cline{2-14} 
 &
   &
  \multicolumn{1}{c|}{CLIP-I} &
  \multicolumn{1}{c|}{CLIP-T} &
  SAS &
  \multicolumn{1}{c|}{CLIP-I} &
  \multicolumn{1}{c|}{CLIP-T} &
  SAS &
  \multicolumn{1}{c|}{CLIP-I} &
  \multicolumn{1}{c|}{CLIP-T} &
  SAS &
  \multicolumn{1}{c|}{CLIP-I} &
  \multicolumn{1}{c|}{CLIP-T} &
  SAS \\ \hline
\textbf{Dreambooth\cite{dreambooth}} &
   &
  \multicolumn{1}{c|}{\textbf{0.93}} &
  \multicolumn{1}{c|}{0.50} &
  0.89 &
  \multicolumn{1}{c|}{0.91} &
  \multicolumn{1}{c|}{0.53} &
  0.85 &
  \multicolumn{1}{c|}{\textbf{0.92}} &
  \multicolumn{1}{c|}{0.52} &
  0.83 &
  \multicolumn{1}{c|}{0.92} &
  \multicolumn{1}{c|}{0.53} &
  0.83 \\ \hline
 \textbf{Textual Inversion~\cite{textual-inversion}} &
   &
  \multicolumn{1}{c|}{0.92} &
  \multicolumn{1}{c|}{\textbf{0.75}} &
  0.83 &
  \multicolumn{1}{c|}{0.90} &
  \multicolumn{1}{c|}{0.74} &
  0.81 &
  \multicolumn{1}{c|}{0.91} &
  \multicolumn{1}{c|}{0.74} &
  0.78 &
  \multicolumn{1}{c|}{0.92} &
  \multicolumn{1}{c|}{0.75} &
  0.79 \\ \hline
\textbf{Custom Diffusion~\cite{custom-diffusion}} &
   &
  \multicolumn{1}{c|}{0.91} &
  \multicolumn{1}{c|}{0.51} &
  0.81 &
  \multicolumn{1}{c|}{0.90} &
  \multicolumn{1}{c|}{0.52} &
  0.79 &
  \multicolumn{1}{c|}{0.91} &
  \multicolumn{1}{c|}{0.51} &
  0.78 &
  \multicolumn{1}{c|}{0.91} &
  \multicolumn{1}{c|}{0.51} &
  0.79 \\ \hline
\multirow{4}{*}{\textbf{Ours w/ scaffolding}} &
  GD=F, SAL=F &
  \multicolumn{1}{c|}{\textbf{0.93}} &
  \multicolumn{1}{c|}{0.72} &
  0.89 &
  \multicolumn{1}{c|}{0.91} &
  \multicolumn{1}{c|}{0.77} &
  0.87 &
  \multicolumn{1}{c|}{0.91} &
  \multicolumn{1}{c|}{0.73} &
  \textbf{0.86} &
  \multicolumn{1}{c|}{0.88} &
  \multicolumn{1}{c|}{0.73} &
  0.81 \\ \cline{2-14} 
 &
  GD=T, SAL=F &
  \multicolumn{1}{c|}{0.92} &
  \multicolumn{1}{c|}{0.72} &
  0.89 &
  \multicolumn{1}{c|}{0.91} &
  \multicolumn{1}{c|}{0.75} &
  0.87 &
  \multicolumn{1}{c|}{0.90} &
  \multicolumn{1}{c|}{0.70} &
  0.81 &
  \multicolumn{1}{c|}{0.87} &
  \multicolumn{1}{c|}{0.69} &
  0.79 \\ \cline{2-14} 
 &
  GD=F, SAL=T &
  \multicolumn{1}{c|}{0.92} &
  \multicolumn{1}{c|}{0.73} &
  0.88 &
  \multicolumn{1}{c|}{0.91} &
  \multicolumn{1}{c|}{0.76} &
  0.87 &
  \multicolumn{1}{c|}{0.91} &
  \multicolumn{1}{c|}{0.73} &
  \textbf{0.86} &
  \multicolumn{1}{c|}{0.88} &
  \multicolumn{1}{c|}{0.73} &
  0.81 \\ \cline{2-14} 
 &
  GD=T, SAL=T &
  \multicolumn{1}{c|}{0.92} &
  \multicolumn{1}{c|}{0.73} &
  \textbf{0.90} &
  \multicolumn{1}{c|}{0.91} &
  \multicolumn{1}{c|}{0.77} &
  0.87 &
  \multicolumn{1}{c|}{0.90} &
  \multicolumn{1}{c|}{0.74} &
  0.82 &
  \multicolumn{1}{c|}{0.87} &
  \multicolumn{1}{c|}{0.72} &
  0.81 \\ \hline
 \multirow{4}{*}{\textbf{Ours w/o scaffolding}} &
  GD=F, SAL=F &
  \multicolumn{1}{c|}{0.92} &
  \multicolumn{1}{c|}{0.72} &
  0.88 &
  \multicolumn{1}{c|}{\textbf{0.93}} &
  \multicolumn{1}{c|}{0.66} &
  \textbf{0.89} &
  \multicolumn{1}{c|}{\textbf{0.92}} &
  \multicolumn{1}{c|}{0.63} &
  \textbf{0.86} &
  \multicolumn{1}{c|}{\textbf{0.93}} &
  \multicolumn{1}{c|}{0.68} &
  0.83 \\ \cline{2-14} 
 &
  GD=T, SAL=F &
  \multicolumn{1}{c|}{0.92} &
  \multicolumn{1}{c|}{0.72} &
  0.88 &
  \multicolumn{1}{c|}{\textbf{0.93}} &
  \multicolumn{1}{c|}{0.65} &
  \textbf{0.89} &
  \multicolumn{1}{c|}{0.91} &
  \multicolumn{1}{c|}{0.63} &
  \textbf{0.86} &
  \multicolumn{1}{c|}{\textbf{0.93}} &
  \multicolumn{1}{c|}{0.68} &
  \textbf{0.84} \\ \cline{2-14} 
 &
  GD=F, SAL=T &
  \multicolumn{1}{c|}{0.92} &
  \multicolumn{1}{c|}{0.73} &
  0.89 &
  \multicolumn{1}{c|}{\textbf{0.93}} &
  \multicolumn{1}{c|}{0.65} &
  \textbf{0.89} &
  \multicolumn{1}{c|}{\textbf{0.92}} &
  \multicolumn{1}{c|}{0.63} &
  \textbf{0.86} &
  \multicolumn{1}{c|}{\textbf{0.93}} &
  \multicolumn{1}{c|}{0.68} &
  \textbf{0.84} \\ \cline{2-14} 
 &
  GD=T, SAL=T &
  \multicolumn{1}{c|}{0.92} &
  \multicolumn{1}{c|}{0.73} &
  0.88 &
  \multicolumn{1}{c|}{0.92} &
  \multicolumn{1}{c|}{0.66} &
  \textbf{0.89} &
  \multicolumn{1}{c|}{0.85} &
  \multicolumn{1}{c|}{0.63} &
  0.82 &
  \multicolumn{1}{c|}{\textbf{0.93}} &
  \multicolumn{1}{c|}{0.68} &
  0.83 \\ \hline
\textbf{Ours w/o scaffolding w/ control-net} &
  \multicolumn{1}{c|}{GD=T, SAL=F} &
  \multicolumn{1}{c|}{0.90} &
  \multicolumn{1}{c|}{0.72} &
  \multicolumn{1}{c|}{0.80} &
  \multicolumn{1}{c|}{0.90} &
  \multicolumn{1}{c|}{\textbf{0.82}} &
  \multicolumn{1}{c|}{0.80} &
  \multicolumn{1}{c|}{0.89} &
  \multicolumn{1}{c|}{\textbf{0.82}} &
  \multicolumn{1}{c|}{0.79} &
  \multicolumn{1}{c|}{0.90} &
  \multicolumn{1}{c|}{\textbf{0.83}} &
  \multicolumn{1}{c|}{0.80} \\ \hline
\end{tabular}%
}
\label{tab:comp-tab}
\end{table}

\vspace{-4mm}
\noindent \textbf{4. Conclusion}  
We introduced MSP as an innovative approach for generating high-quality images featuring multiple personalized subjects. Our experimental results highlight the superior performance of our method compared to text-to-image models such as Textual Inversion, Dreambooth and Custom-Diffusion.

\noindent \textbf{Ethical Implications}  
Text-to-image generation carries substantial ethical considerations. While these models offer valuable applications like creative content generation and marketing, they also open the door for misuse, enabling the creation of deceptive images. Another thing to mention here is that our evaluation dataset utilizes target subjects from Comicgen~\footnote{Comicgen: \url{https://gramener.com/comicgen/}} and Alamy~\footnote{Alamy \url{https://www.alamy.com/}} strictly for research purposes.

\bibliographystyle{ACM-reference-format}
\bibliography{main.bib}

\appendix
\section{Appendix}
\label{ap-sec:appendix}
We provide supplementary material to the paper in this Appendix. Herein we describe the sample dataset (\ref{ap-sec:dataset}), point out the limitations of the existing methods for personalizing multiple subjects (\ref{ap-sec:problem}), provide algorithmic and implementation details of our methods(\ref{ap-sec:input-format}, \ref{ap-sec:sal}, \ref{ap-sec:exp-setup}), provide evaluation metrics(\ref{ap-sec:eval-metric}), discuss more qualitative results (\ref{ap-sec:results}) and discuss the limitations (\ref{ap-sec:limitations}) and key contributions of our work (\ref{ap-sec:key_contributions}).

\subsection{Dataset}
\label{ap-sec:dataset}
\begin{figure*}[!hbtp]
  \centering
  \includegraphics[width=\textwidth]{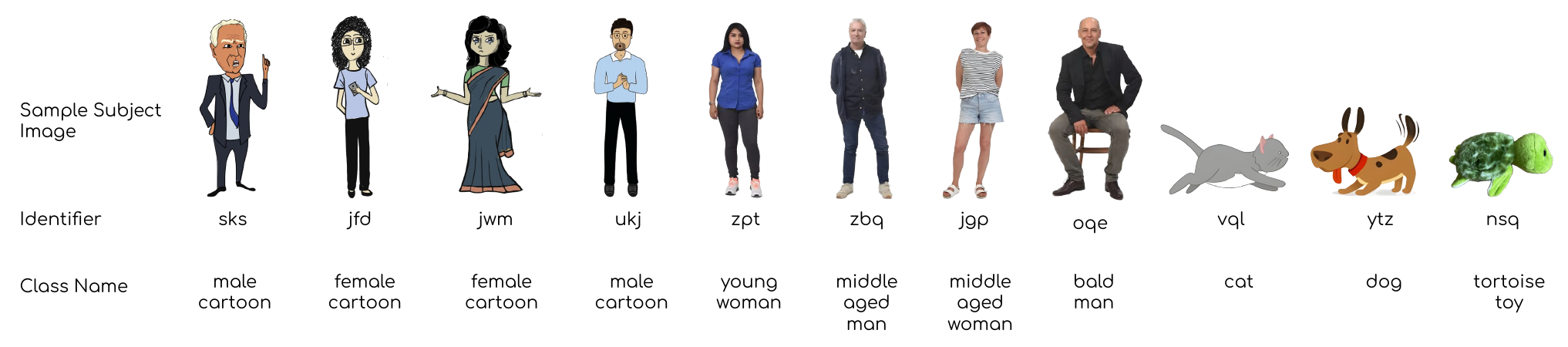}
  \caption{Figure displaying sample training images, each linked to a distinct subject, alongside the unique identifier and class name used in the fine-tuning process of the Dreambooth model.}
  \label{fig:training-set}
\end{figure*}

We assess MSP-Diffusion on our self-compiled Subject-Dataset11\footnote{Subject-Dataset11: \url{https://tinyurl.com/2j7jbzzx}} 
featuring $11$ diverse target subjects, including comic characters (male, female), human characters (male, female), pets, and still objects. We obtain $11$ training images corresponding to each target subject from online sources such as Comicgen~\footnote{Comicgen: \url{https://gramener.com/comicgen/}} and Alamy~\footnote{Alamy: \url{https://www.alamy.com/}}. Below, we present sample target images of the 11 subjects featured in our Subject-Dataset11 which we plan to release publicly strictly for research purposes. In Figure~\ref{fig:training-set}, you can observe the training images, each accompanied by unique identifiers and the corresponding subject class, used for fine-tuning the Dreambooth model. For instance, "sks male cartoon" features "sks" as the unique identifier and "male cartoon" as the class to which the subject belongs. We employed 11 different instances of each target subject image, featuring variations in pose (sitting, standing, walking) and orientation (front, back, left, right). These images were instrumental in fine-tuning the Dreambooth model. 

Given the limited number of training images per target subject for fine-tuning Dreambooth and the UNet segmentation model, we augment the dataset with transformations. This includes zooming subjects (scaling factors 0.4 to 0.5), cropping subjects (0.65 to 0.95 of the original size with 0.1 probability), and introducing random translations on a $512 \times 512$ image canvas. This augmentation results in a dataset of 1064 training images, 194 validation images, and 150 test images. To further enhance the UNet segmentation model's effectiveness in isolating target subjects for subject-aware loss and improved subject appearance during image generation, we introduce 100 diverse background scenes to these images. The process of cropping and merging backgrounds is illustrated in Figure~\ref{fig:msp-diffusion-architecture}.

The evaluation dataset includes 11 different compositions of the target subjects in various settings (1-subject, 2-subjects, 3-subjects, and >3 subjects) with 5 text prompts for each composition. We generate 4 images per text prompt and report the average evaluation metrics.

\subsection{Problem with existing methods}
\label{ap-sec:problem}
\begin{figure*}[!htbp]
  \centering
  \includegraphics[width=\textwidth]{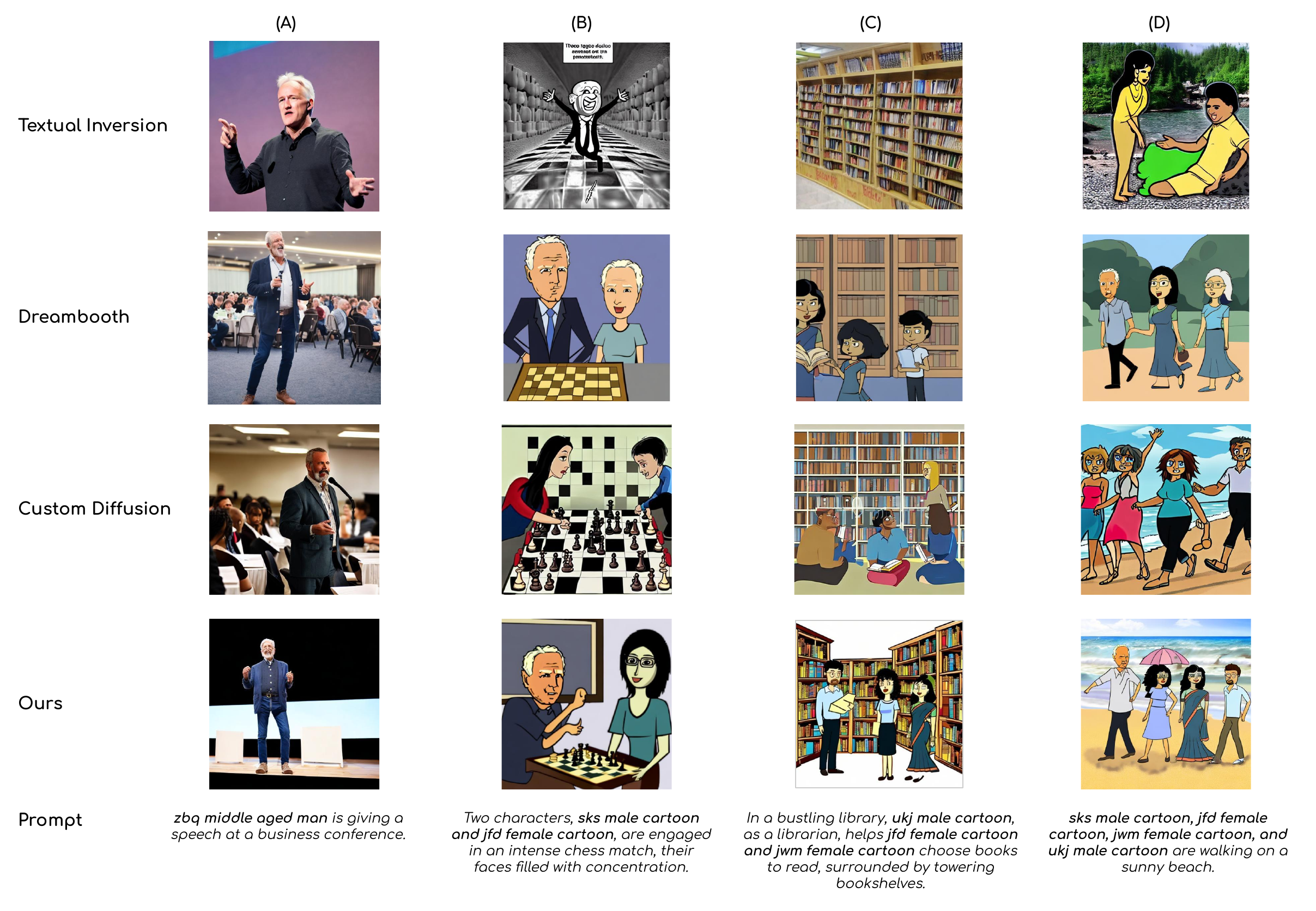}
  \caption{\small{Qualitative comparison of MSP-Diffusion against Dreambooth~\cite{dreambooth}, Textual Inversion~\cite{textual-inversion} and Custom Diffusion~\cite{custom-diffusion}. Columns A, B, C and D refer to the images having number of personalized subjects as 1, 2, 3 and more than 3, respectively. Here, it is clearly visible that models like Dreambooth, Textual-Inversion and Custom Diffusion struggle to generate images with multiple personalized subjects and produce poor quality images (missing subjects, hybrid subject exhibiting the characteristics of multiple subjects and wrong subject's appearance).}}
  \label{fig:comp-results}
\end{figure*}

Figure~\ref{fig:comp-results} showcases image quality comparisons between previous methods (Textual-Inversion~\cite{textual-inversion}, Dreambooth~\cite{dreambooth} and Custom-Diffusion~\cite{custom-diffusion}) and our MSP-Diffusion approach. Columns (A), (B), (C), and (D) represent image generation scenarios with 1, 2, 3, and more than 3 personalized subjects. For single subject personalization, previous methods perform well. However, as the number of personalized subjects increases, issues such as missing subjects, incorrect appearances, and hybrid subjects with mixed characteristics become apparent, as seen in columns (B), (C), and (D) of rows 1, 2 and 3 in Figure~\ref{fig:comp-results} for Textual-Inversion, Dreambooth and Custom-Diffusion.

Therefore, our primary goal is to accurately generate images with multiple personalized subjects based on a given text prompt. In Row 4 of Figure~\ref{fig:comp-results}, we present the results of our proposed MSP-Diffusion method, where we have successfully addressed the limitations observed in previous approaches.

\begin{figure*}[!hbtp]
  \centering
  \includegraphics[width=\textwidth]{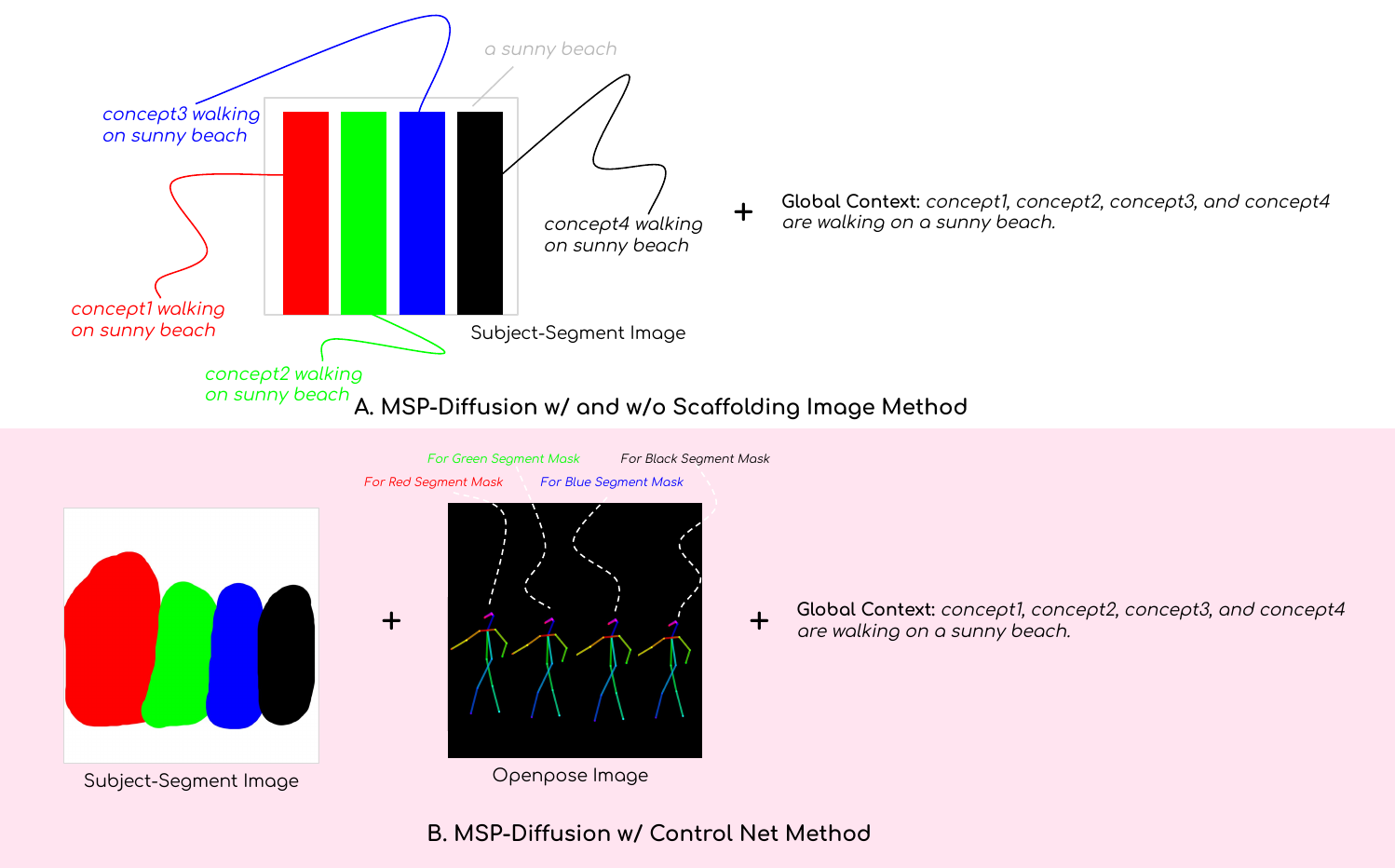}
  \caption{Figure illustrates a sample input to the MSP-Diffusion method during inference. (A) presents the input format for scenarios both with and without a scaffolding image, while (B) provides additional details regarding an Openpose image in the context of ControlNet conditioning with MSP-Diffusion. It's worth noting that \textit{conceptI} can be substituted with the specific subject, such as "sks male cartoon," for instance.}
  \label{fig:input-sample}
\end{figure*}

\subsection{Input Format of MSP-Diffusion}
\label{ap-sec:input-format}
Figure~\ref{fig:input-sample} depicts the MSP-Diffusion inference input format. It consists of a colored subject-segment image with segments representing different personalized subjects, each accompanied by its respective local text prompt. Additionally, a global text prompt is provided, conveying the overall context for the generated image. This global prompt plays a vital role in harmonizing individual segments and capturing the intended subject interactions, aligning with the specifications in the text prompt. When we incorporate ControlNet's~\cite{control-net} openpose~\cite{openpose} into the MSP-Diffusion method, we also provide an additional Openpose image, as shown in Figure~\ref{fig:input-sample}(B) describing the posture and size of the personalized subjects.

\subsection{Subject-Aware Segmentation Loss (SAL)}
\label{ap-sec:sal}
Given an input image denoted as ${x_0}$, an image diffusion algorithm progressively introduces noise to the image, resulting in the creation of a noisy image referred to as $x_t$. Here, the variable $t$ signifies the number of iterations at which noise is incrementally added. At each time step $t$, this image diffusion algorithm employs a neural network to acquire the ability to predict the noise components $\epsilon_{u}$ (unconditional) and $\epsilon_{c}$ (conditional) that are applied to the noisy image $x_t$. This predictive process encompasses two distinct scenarios: one where no specific conditions are imposed, and the other involving the incorporation of particular conditions conveyed through textual prompts denoted as $c$. The forward process variances $\beta_{t}$ can be learned by re-parameterization. Considering, reverse diffusion process for an input noisy image $x_{t}$, $\epsilon_{u}$ as unconditional noise prediction and $\epsilon_c$ as conditional noise prediction, and $\alpha_{t} := 1-\beta_{t}$ ~\cite{ddpm}, the final projected image $\hat{x_0}$ at time $t$ is given by:
\begin{displaymath}
\hat{x_0}\leftarrow \frac{x_t}{\sqrt{\alpha _t}}- \frac{\sqrt{1-\alpha_t}}{\sqrt{\alpha_t}} \epsilon
\end{displaymath}

\begin{displaymath}
where, \epsilon = \epsilon_u + 7.5 * (\epsilon_c - \epsilon_u)
\end{displaymath}

Since, we do not have the ground truth annotation for the target subject class, we use direct entropy minimization~\cite{advent} to accentuate the most likely regions of the subject. Given segment-specific region $m^{seg_j}$ for segment $j$, we define $x_{0,j}$ as $\hat{x_0}\odot m^{seg_j}$ which is given as input to the trained subject-aware segmentation model ($M_{SAL}$). The output of the segmentation is given below:
\begin{displaymath}
    \mathbf{P_{x_{0,j}}}^{(h,w,c)} = {M_{SAL}(\hat{x_0}\odot m^{seg_j})}
\end{displaymath}

Next, we compute the Entropy map $\mathbf{E_{x_{0,j}}} \in [0,1]^{H*W}$ which is composed of the normalized pixel-wise entropies as follows:


\begin{displaymath}
    \mathbf{E _{x_{0,j}}}^{(h,w)} = - \frac{1}{\log{(C)}} \sum_c{\mathbf{{P_{x_{0,j}}}}^{(h,w,c)} \log \mathbf{{P_{x _{0,j}}}}^{(h,w,c)}}
\end{displaymath}

where, $c$ is the target subject-class obtained from the local text-prompt. Subsequently, the entropy loss is defined as the average of all pixel wise normalized entropies:

\begin{displaymath}
    \mathcal{L}_{ent}({x_{0,j}}) = \frac{1}{h*w} \sum\limits_{h,w} \mathbf{E _{x_{0,j}}}^{(h,w)}
\end{displaymath}

Finally, we provide external guidance to the diffusion model by computing gradient from the entropy loss as follows:
\begin{displaymath}
    \triangledown ^{seg_j}\leftarrow \triangledown \mathcal{L}_{ent}({x_{0,j}})
\end{displaymath}





\begin{figure*}[!htbp]
  \centering
  \includegraphics[width=\textwidth]{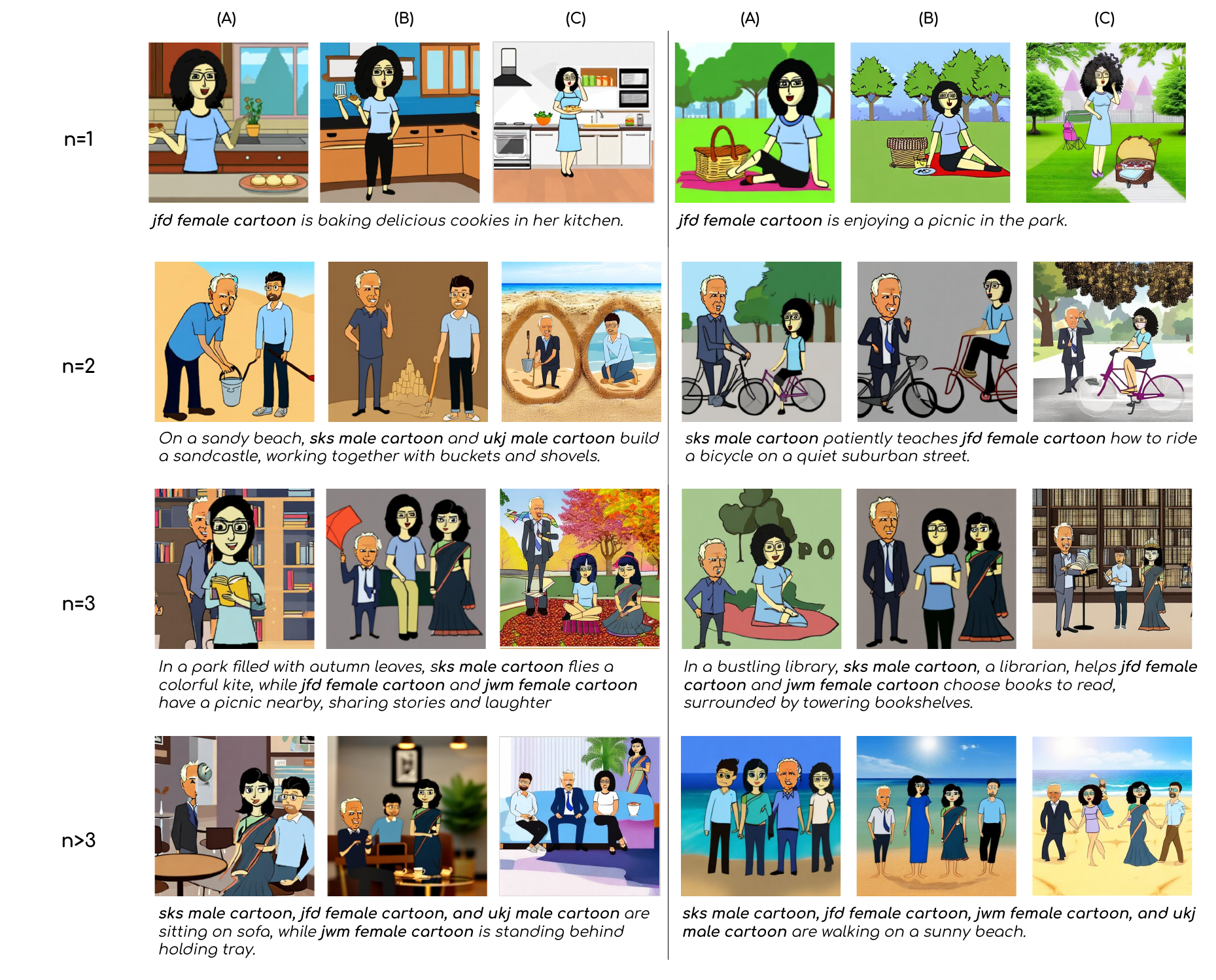}
  \caption{Figure showing the qualitative results of proposed MSP-Diffusion model in settings such as (A) with scaffolding image, (B) without scaffolding image and (C) with  ControlNet conditioning. Here, $n$ refers to the number of personalized subjects in the image-generation.}
  \label{fig:ablation-figure}
\end{figure*}

\subsection{Experimental Setup}
\label{ap-sec:exp-setup}
The Dreambooth model is fine-tuned on a T4 12GB GPU, taking 4 hours with a batch size of 1 for a total of 22,000 training steps. Additionally, the ResNet-34 based UNet for subject segmentation was trained on T4 GPU with a batch size of $4$ and learning rate of $1e-4$ for $14$ epochs. In the case of MSP-Diffusion, 80 diffusion steps were employed for two settings (with and without scaffolding image), while only 50 diffusion steps were used with ControlNet. MSP-Diffusion with a scaffolding image initially utilized the global diffuser for 8 steps, whereas MSP-Diffusion without scaffolding and with ControlNet required only 2 global diffuser steps. The inference for MSP-Diffusion (with and without scaffolding image) was performed on a 10GB MIG A100 GPU. Conversely, image generation via MSP-Diffusion with ControlNet was accomplished using Google Colab Pro, leveraging 12 GB T4 GPU.

\subsection{Evaluation Metrics}
\label{ap-sec:eval-metric}

We compare the performance of MSP-Diffusion against existing text-to-image generation methods such as Textual-Inversion~\cite{textual-inversion}, Dreambooth~\cite{dreambooth} and Custom-Diffusion~\cite{custom-diffusion} based on two essential criteria: subject similarity and context coverage, employing three distinct assessment methods as follows:
\begin{itemize}
\item \textbf{CLIP-I~\cite{textual-inversion}}: It quantifies similarity by considering semantic distances in the CLIP image-space. Specifically, CLIP-I calculates the cosine similarity between the CLIP embeddings of generated images and their corresponding subject-specific source images.

\item \textbf{Subject-aware similarity (SAS)}: We found that the CLIP scoring mechanism is not inherently aware of the target subjects in the embedding space. Consequently, even quite dissimilar subjects may yield a CLIP-I score of $0.8$. To address this limitation, we propose an alternative evaluation method called Subject-aware similarity (SAS). It computes cosine similarity between subject-aware feature embeddings, leveraging the latent features extracted from the encoder of the Subject-aware UNet segmentation model. 

\item \textbf{CLIP-T ~\cite{text-alignment}}: We measure context similarity using CLIP-Text Similarity, computed as the average similarity score for the best generations corresponding to each individual text-prompt across all compositions in the CLIP-space.
\end{itemize}

\begin{figure*}[t]
  \centering
  \includegraphics[width=\textwidth]{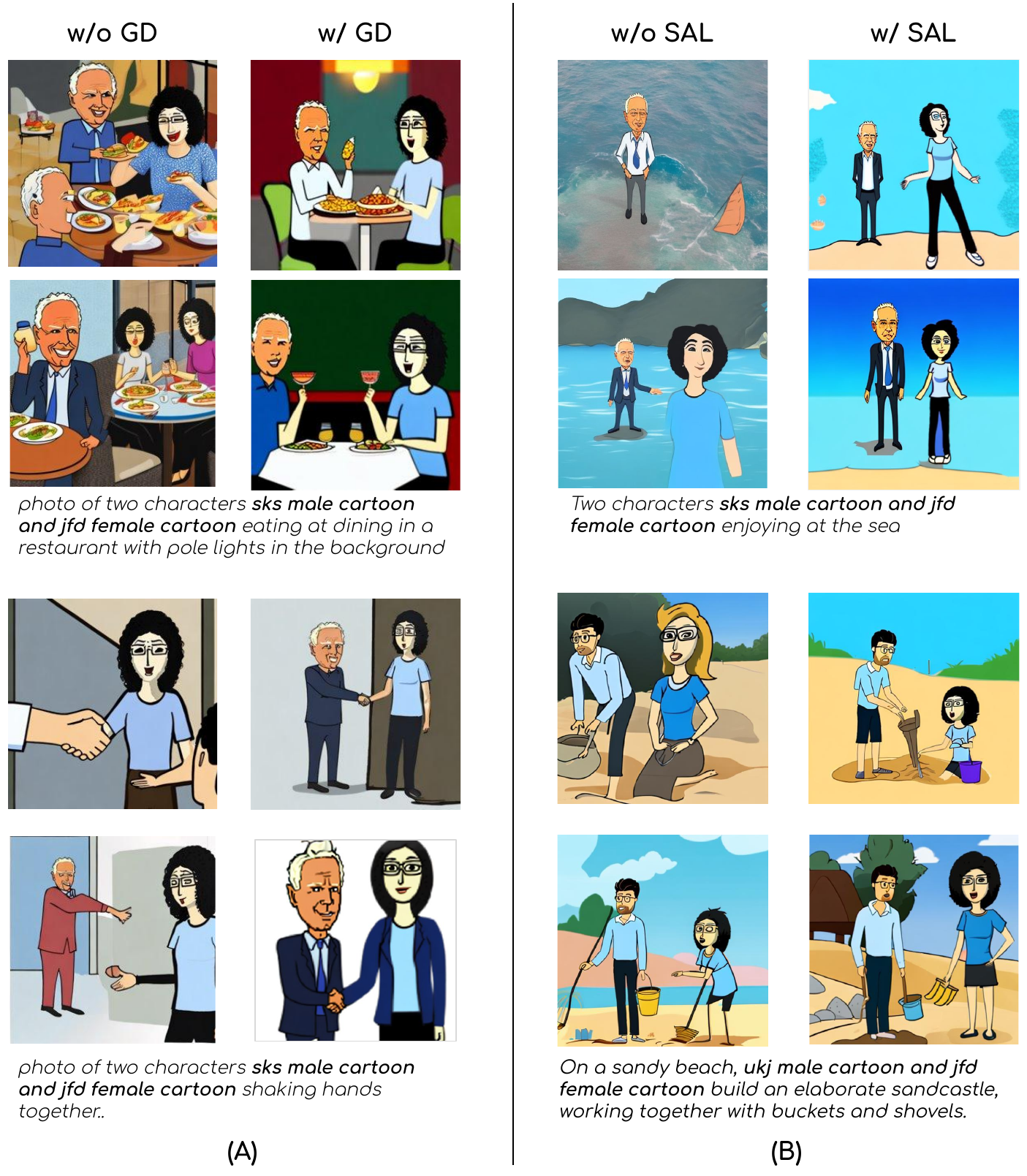}
  \caption{Ablation study on adding Global Diffuser (GD) and Subject-aware Loss (SAL) in MSP-Diffusion. (A) demonstrates that the generated image becomes more harmonized with the integration of GD, (B) illustrates that the subject's appearance in the generation looks more like the corresponding training images with the integration of SAL.}
  \label{fig:gd-and-sas}
\end{figure*}

\begin{figure*}[!hbtp]
  \centering
  \includegraphics[width=0.9\textwidth]{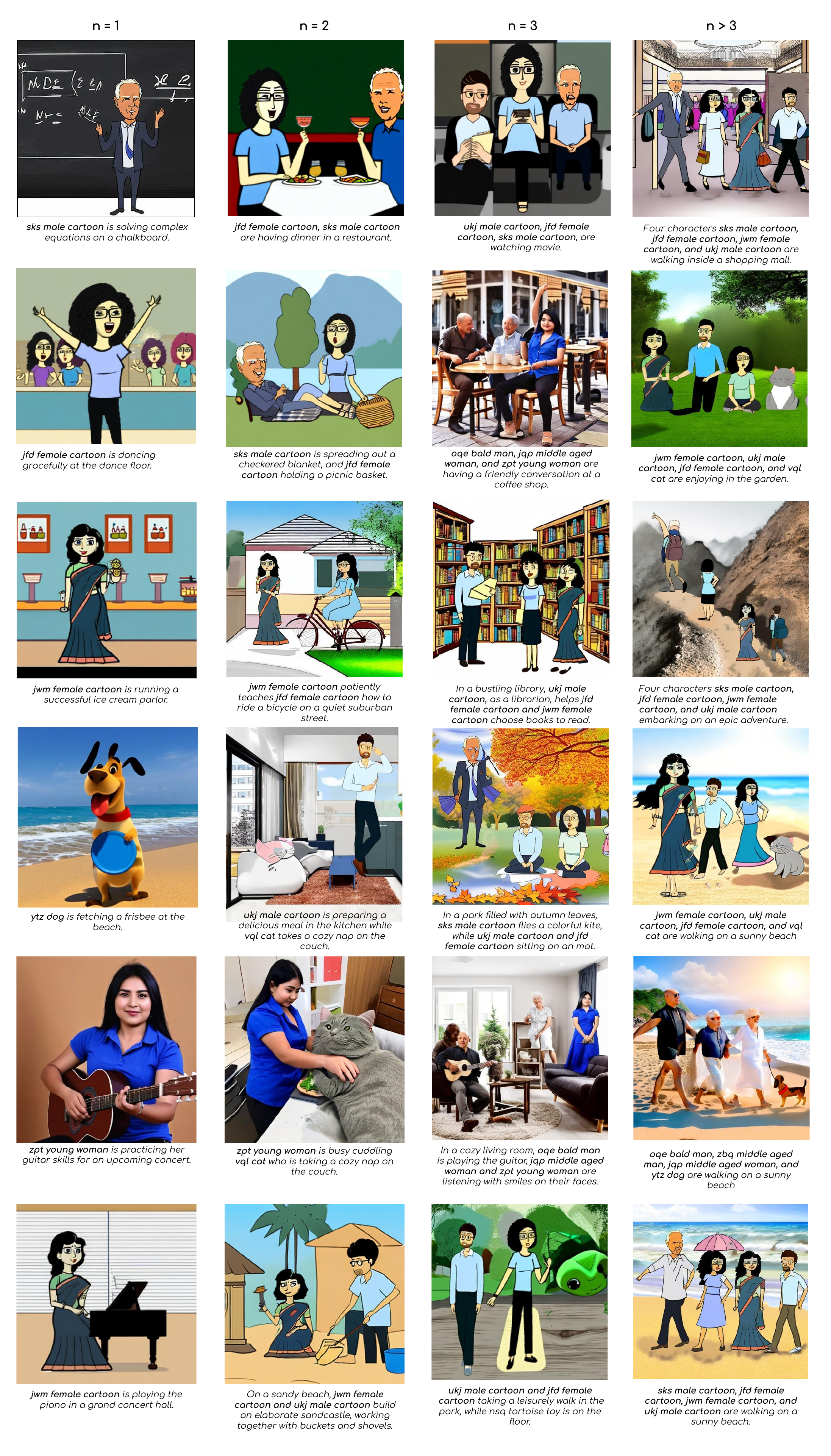}
  \vspace{-2mm}
  \caption{Qualitative results of our proposed MSP-Diffusion model for generating images featuring multiple personalized subjects. Here $n$ represents the number of personalized subjects in the generated image.}
  \label{fig:msp-results}
\end{figure*}

\subsection{Qualitative Results}
\label{ap-sec:results}
Here, we present a comprehensive qualitative analysis of the MSP-Diffusion method across three different settings: (A) with a scaffolding image, (B) without a scaffolding image, and (C) with ControlNet's Openpose control-conditioning. Figure~\ref{fig:ablation-figure} provides a visual representation of the performance in generating multiple personalized subjects, clearly demonstrating the effectiveness of our proposed method.
In the case of MSP-Diffusion without a scaffolding image, although the background is occasionally missing, the number and appearance of personalized subjects surpass the method with a scaffolding image. Conversely, MSP-Diffusion with the incorporation of ControlNet conditioning yields the best results. However, it's important to note that this method is computationally intensive and relies on an additional dependency, requiring an Openpose image as input during inference.

Additionally, we present a qualitative ablation analysis that underscores the effectiveness of integrating a Global Diffuser and Subject-aware Loss (SAL) into the MSP-Diffusion method. In Figure~\ref{fig:gd-and-sas}(A), it is evident that without the inclusion of a global diffuser, the interactions and overall context of the generated image lack harmony. However, upon incorporating the global diffuser, the personalized subjects in the generated images exhibit harmonized interactions and a consistent background. 

Moreover, as depicted in Figure~\ref{fig:gd-and-sas}(B), it becomes evident that the appearance of personalized subjects was inaccurate without the inclusion of SAL, resulting in subjects that differed from their corresponding training images. To maintain control over the intended subject's appearance, we introduced SAL, ensuring that the subject's appearance was faithfully retained.

Now, we provide some more qualitative results of MSP-diffusion for generating images with multiple personalized subjects in Figure~\ref{fig:msp-results}. This showcases that we are able to generate good quality harmonized images having multiple personalized subjects with the intended interactions between subjects.

\subsection{Limitations}
\label{ap-sec:limitations}
Our research paper, titled "MSP-Diffusion," presents compelling findings, as illustrated in Figure~\ref{fig:msp-results}. Upon closer examination of the generated content, it is evident that the quality of generations involving distinct subject categories, such as comics featuring real human characters, falls short of our expectations. We are actively engaged in ongoing efforts to enhance this aspect of our work.

Furthermore, we have observed occasional instances where the background scene in the image generated using MSP-Diffusion without a scaffolding image appears to be missing. We are currently working on resolving this issue. Additionally, we are committed to reducing our reliance on Openpose for the generation process, exploring alternatives to further improve our methodology.

\subsection{Key Contributions}
\label{ap-sec:key_contributions}

Our paper contributes in the following ways:

\vspace{-2mm}
\begin{enumerate}

    \item A novel multi-subject aware personalized text-to-image generation approach named MSP-Diffusion, leveraging Stable Diffusion techniques to consistently produce high-quality images featuring multiple personalized subjects.
    \item We utilize a limited number of training samples per personalized target subject, facilitating the fine-tuning of the base Dreambooth model.
    \item We employ Composite Diffusion~\cite{composite-diffusion} alongside our subject-aware loss for personalized subjects within an image segment, allowing control over subject appearance.
    \item We introduce a global diffuser capturing the global context of the image to ensure harmonization among various image segments containing personalized subjects.
    \item A comprehensive comparison with existing state-of-the-art text-to-image generation models on a self-curated dataset of textual descriptions involving multiple personalized subjects. 
    \
\end{enumerate}

\end{document}